\documentclass[conference]{IEEEtran}
\usepackage{cite}
\usepackage{amsmath,amssymb,amsfonts}
\usepackage{algorithmic}
\usepackage{graphicx}
\usepackage{epstopdf}
\usepackage{subcaption}
\usepackage{textcomp}
\usepackage{xcolor}
\usepackage{bm}
\usepackage{bbm}
\usepackage{caption}
\usepackage[ruled,lined]{algorithm2e}
\usepackage{algorithmic}
\usepackage{blindtext}
\usepackage{soul}
\usepackage{subfloat} 

\usepackage{mathtools}

\graphicspath{{Fig/},{fig/}}

\begin{document}

\newcommand{\fig}{Fig.\xspace}

\title{\huge Multi-Agent Deep Reinforcement Learning for Energy Efficient Multi-Hop STAR-RIS-Assisted Transmissions
}

\author{\IEEEauthorblockN{Pei-Hsiang Liao, Li-Hsiang Shen$^*$, Po-Chen Wu, and Kai-Ten Feng}
\IEEEauthorblockA{ Department of Electronics and Electrical Engineering, National Yang Ming Chiao Tung University, Hsinchu, Taiwan \\
$^{*}$Department of Communication Engineering, National Central University, Taoyuan, Taiwan \\
Email: phliao.ee11@nycu.edu.tw, shen@ncu.edu.tw, wupochen.ee11@nycu.edu.tw, and ktfeng@nycu.edu.tw}}

\maketitle

%\linespread{0.9}

\begin{abstract}
Simultaneously transmitting and reflecting reconfigurable intelligent surface (STAR-RIS) provides a promising way to expand coverage in wireless communications. However, limitation of single STAR-RIS inspire us to integrate the concept of multi-hop transmissions, as focused on RIS in existing research. Therefore, we propose the novel architecture of multi-hop STAR-RISs to achieve a wider range of full-plane service coverage. In this paper, we intend to solve active beamforming of the base station and passive beamforming of STAR-RISs, aiming for maximizing the energy efficiency constrained by hardware limitation of STAR-RISs. Furthermore, we investigate the impact of the on-off state of STAR-RIS elements on energy efficiency. To tackle the complex problem, a Multi-Agent Global and locAl deep Reinforcement learning (MAGAR) algorithm is designed. The global agent elevates the collaboration among local agents, which focus on individual learning. In numerical results, we observe the significant improvement of MAGAR compared to the other benchmarks, including Q-learning, multi-agent deep Q network (DQN) with golbal reward, and multi-agent DQN with local rewards. Moreover, the proposed architecture of multi-hop STAR-RISs achieves the highest energy efficiency compared to mode switching based STAR-RISs, conventional RISs and deployment without RISs or STAR-RISs.
\end{abstract}


\footnotetext[1]{This work was supported in part by the National Science and Technology Council (NSTC) under Grant NSTC 113-2221-E-A49-119-MY3, Grant 113-2218-E-A49-026, Grant 113-2218-E-A49-027, Grant 112UC2N006, Grant 112UA10019; in part by the Higher Education Sprout Project of the National Yang Ming Chiao Tung University (NYCU) and Ministry of Education (MoE); in part by the Co-creation Platform of the Industry-Academia Innovation School, NYCU, under the framework of the National Key Fields Industry-University Cooperation and Skilled Personnel Training Act, from the MOE and industry partners in Taiwan, and in part by the Hon Hai Research Institute, Taipei, Taiwan.}

\section{Introduction}
Reconfigurable intelligent surfaces (RISs) have emerged as a promising technology to improve the capacity and reliability of wireless communication \cite{RIS_tutorial3}. By controlling and manipulating configurations of RIS elements, a virtual line-of-sight (LoS) link is created to bypass obstacles between transceivers \cite{Lab2_configurations}. 
Leveraging these significant benefits, study \cite{Lab3_RIS} aims to maximize the downlink transmission rate. 
Consequently, RIS technology enables the achievement of targeted signal distributions, leading to significant improvements in communication performance\cite{Lab4}.
However, limited by only reflecting signals, RIS cannot be fully utilized when the transmitters and receivers are located on opposite sides.

As a remedy, simultaneously transmitting and reflecting RIS (STAR-RIS) enables 360$^\circ$ full-plane coverage. In contrast to traditional RIS, the STAR-RIS eliminates deployment restrictions to specific geographic areas. A growing body of research investigates the use of STAR-RIS to enhance communication systems \cite{STAR-RIS_design1,STAR-RIS_design2}. In \cite{STAR-RIS_design1}, the authors analyze the energy efficiency of the system with optimization of the beamforming of BS and the configuration of STAR-RIS. In addition, the authors of \cite{STAR-RIS_design2} integrate the mode switching design in STAR-RIS to maximize the achievable sum-rate. 
Although research into STAR-RIS intensifies, the potential of cooperative multiple STAR-RISs is emerging as an alternative approach to enhance coverage.

Since multiple distributed RISs without signal cooperation uniquely reflects the signals to users, the concept of multi-hop transmissions has been proposed to enlarge the coverage and connectivity. Furthermore, multi-hop reflections are established through inter-RIS links, and provide blockage-free links with higher transmission quality \cite{multi-hop1,multi-hop3}. The authors of \cite{multi-hop1} propose a machine learning-based approach to optimize RIS beamforming. In \cite{multi-hop3}, the authors design the beamforming of BS and RISs by leveraging deep learning. However, existing research has primarily focused on multi-hop RIS systems. Inspired by these approaches, we propose the novel architecture of multi-hop STAR-RISs, enabling a wider range of full-plane service coverage.

In this paper, we present a multi-hop communications assisted by STAR-RISs to overcome the propagation attenuations and improve the coverage range. In particular, we consider the on-off state of STAR-RIS elements, which is not considered in most of existing papers. However, traditional algorithms possess high complexity to find solutions with large coefficients. Deep reinforcement learning (DRL) is especially suitable for unknown environments without channel estimation\cite{DRL-JCCRA, DRL-RIS, DRL-RIS_FD}. To achieve long-term performance benefits in a multi-hop STAR-RIS system, we design a DRL-enhanced algorithm to optimize joint active BS and passive STAR-RIS beamforming. The contributions of this work are summarized as follows.

\begin{itemize}
    \item We consider multi-hop STAR-RISs with joint optimization of active BS and passive STAR-RIS beamforming for maximizing the system energy efficiency. Additionally, we investigate the on-off mechanism of STAR-RIS elements to address the potential of high power consumption of STAR-RISs.

    \item We have proposed a Multi-Agent Global and locAl deep Reinforcement learning (MAGAR) algorithm with collaborative multiple agents. Local agents perform local learning and facilitate independent interaction with the environment. In contrast, the global agent periodically replaces local agents to dictate actions, thereby optimizing the overall system performance.

    \item Benefiting from the global-local agent design, MAGAR achieves the highest energy efficiency among the other benchmarks of multi-agent deep Q network (DQN) and conventional Q-learning. Moreover, the multi-hop STAR-RISs have achieved higher performance than the mode switching based STAR-RISs, conventional RISs, and deployment without RISs/STAR-RISs.
\end{itemize}
\section{System Model and Problem Formulation}
\begin{figure}
 \centering
 \includegraphics[width=3.3in]{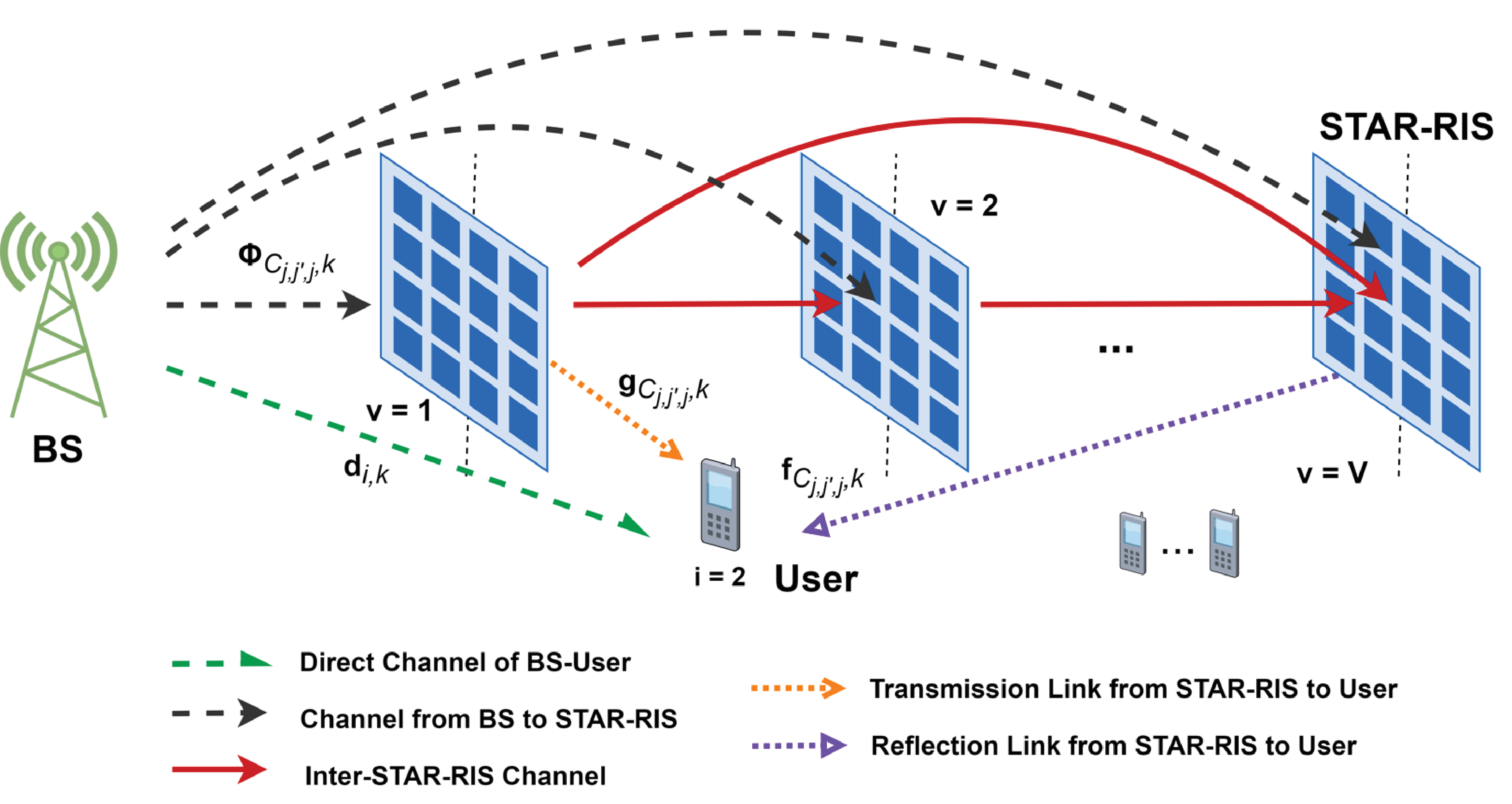} % change to eps file when this paper finish !
 \captionsetup{font=footnotesize}
 \caption{System architecture of the proposed multi-hop STAR-RIS}
 \label{Fig:systemModel}
\end{figure}
\subsection{System Model}
As depicted in Fig. \ref{Fig:systemModel}, we consider a single BS serving $K_{i}$ user equipments (UEs) per region, with $I$ total regions. Note that all users are equipped with a single antenna. The $V$ STAR-RIS placed to enhance signal propagation and potentially enhance service coverage.
The respective sets of STAR-RISs, service region and users are defined as $\mathcal{V} = \left\{1, ..., V \right\}$, $\mathcal{I} = \left\{1, ..., I \right\}$ and $\mathcal{K}_{i} = \left\{1, ..., K_{i} \right\}$.
The propagation of signals is also delineated in Fig. \ref{Fig:systemModel}, illustrating all conceivable signal pathways, assuming a user is located in $2$-th region.
We define $\mathcal{C} = \left\{\mathcal{C}_{j}, \forall j \in \mathcal{V}\right\}$ as the set corresponding to all possible paths, where $j$ indicates the number of STAR-RISs which the signal experiences through. Furthermore, $\mathcal{C}_{j} = \left\{{\mathbf{C}}_{j,j'}, \forall {j'} \in \left|\mathcal{C}'_{j}\right|\right\}$ is the set of all possible paths with $j$ STAR-RISs, where the tuple ${\mathbf{C}}_{j,j'} = \left(C_{j,j',1}, ..., C_{j,j',j}\right)$ corresponding to the signal passing through $j$ STAR-RISs.
In each item, the first subscript of each item $j$ is the number of signal passing through, the second subscript $j'$ is the index of possible permutations of the set $\mathcal{C}'_{j}$ in lexicographic order, and the third subscript is the index of the permutation. 
Moreover, we assume that the orientations of all STAR-RISs face forward the BS. Note that the property of each item in ${\mathbf{C}}_{j,j'}$ represents the index of STAR-RIS will be arranged in ascending order.
\subsection{STAR-RIS}
The BS is equipped with a uniform linear array comprising $M$ transmit antennas and the STAR-RISs we employed are energy splitting (ES)\cite{STAR-RIS_mode}. We define ${N_v}$ as the number of elements of the $v$-th STAR-RIS. For analysis convenience, we consider $N_v = N, \forall v \in \mathcal{V}$ with a set of $\mathcal{N} = \left\{1,\dots,N\right\}$. Furthermore, we denote the set $\mathcal{U} = \left\{\mathcal{R}, \mathcal{T}\right\}$ as the reflection and transmission functionality set, respectively. For the $v$-th STAR-RIS, the on-off matrix ${\boldsymbol{\Lambda}}_{v}$ and the passive beamforming matrix ${\boldsymbol{\Xi}}^{u}_{v}$ having diagonal structures are denoted as
\begingroup
\small
\begin{subequations}
\begin{align}        
    &{\boldsymbol{\Lambda}}_{v} = \text{diag}\left(\alpha_{v, 1},...,
    \alpha_{v, n},...,\alpha_{v, N}\right), 
    \qquad \forall {v} \in \mathcal{V}, \\
    &{\boldsymbol{\Xi}}^{u}_{v} = \text{diag}\left(\beta_{v, 1}^{u}e^{j\theta_{v, 1}^{u}},...,\beta_{v, n}^{u}e^{j\theta_{v, n}^{u}},...,\beta_{v, N}^{u}e^{j\theta_{v, N}^{u}}\right),     \nonumber\\
    &\qquad \qquad \qquad \qquad \qquad \qquad \qquad \qquad \
    \forall {v} \in \mathcal{V}, {u} \in \mathcal{U}, \\
    \label{eqn:theta_matrix} 
    &{\boldsymbol{\Theta}}^{u}_{v} = {\boldsymbol{\Lambda}}_{v}{\boldsymbol{\Xi}}^{u}_{v}, \qquad \forall {v} \in \mathcal{V}, {u} \in \mathcal{U},
\end{align}
\end{subequations}
\endgroup

where $\alpha_{v, n} \in \left\{0, 1\right\}$ stands for whether the $n$-th element of STAR-RIS is selected to be switched on. The notations of ${\beta_{v, n}^{u}} \in \left[0, 1\right]$ and ${\theta_{v, n}^{u}} \in \left[0, 2\pi\right)$ represent the amplitude and phase shift of the $n$-th element of STAR-RIS, respectively. We represent the set of on-off as ${\boldsymbol{\Lambda}} = \left\{{\boldsymbol{\Lambda}}_{v}, \forall {v} \in \mathcal{V}\right\}$ and passive beamforming ${\boldsymbol{\Xi}} = \left\{{\boldsymbol{\Xi}}_{v}^{u}, \forall {v} \in \mathcal{V}, {u} \in \mathcal{U}\right\}$. Furthermore, due to limitations imposed by the electric and magnetic impedance, the energy conservation constraint and the coupling effect between the reflected and transmitted phase shifts \cite{STAR-RIS_constraints} is obtained as
\begingroup
\begin{subequations} \label{constraint}
\begin{align}
    &\beta_{v, n}^{\mathcal{T}} = \sqrt{1-\left(\beta_{v, n}^{\mathcal{R}}\right)^2} ,\qquad \forall {v} \in \mathcal{V}, {n} \in \mathcal{N},\\
    &\theta_{v, n}^{\mathcal{T}} = 
        \theta_{v, n}^{\mathcal{R}} \pm \frac{\pi}{2},
    \qquad \qquad \forall {v} \in \mathcal{V}, {n} \in \mathcal{N}.
\end{align}
\end{subequations}
\endgroup

\subsection{Channel Model}
For the $k$-th user in the region $i$, we denote ${\rm\mathbf{d}}_{i, k} \in \mathbb{C}^{M \times 1}$, ${\rm\mathbf{g}}_{C_{j,j',j}, k} \in \mathbb{C}^{N \times 1}$, and ${\rm\mathbf{f}}_{C_{j,j',j}, k} \in \mathbb{C}^{N \times 1}$ as direct channels from the BS to the user, transmitted channels, and reflected channels from the ${C_{j,j',j}}$-th STAR-RIS to the user, respectively. The channels from the BS to the ${C_{j,j',1}}$-th STAR-RIS is denoted as ${\rm\mathbf{\Phi}}_{ C_{j,j',1}} \in \mathbb{C}^{M \times N}$ and channels from the ${C_{j,j',l}}$-th to the ${C_{j,j',l+1}}$-th STAR-RISs are denoted as ${\rm\mathbf{\Phi}}_{C_{j,j',l},C_{j,j',l+1}} \in \mathbb{C}^{N \times N}$. 
We assume that the channels associated with the BS and STAR-RISs is indicated by ${\rm\mathbf{H}} \in \left\{{\rm\mathbf{d}}_{i, k}, {\rm\mathbf{g}}_{C_{j,j',j}, k}, {\rm\mathbf{f}}_{C_{j,j',j}, k}, {\rm\mathbf{\Phi}}_{C_{j,j',l},C_{j,j',l+1}}\right\}$ obeys the Rician distribution which is given by
%, since the selected positions of the BS and STAR-RISs provide a line-of-sight (LoS) path\cite{channel_fading}. Thus, considering the large scale \cite{pathloss_reference} of pathloss during signal propagation, the Rician fading channel has been formulated as
\begingroup
\allowdisplaybreaks
\begin{equation}
    \begin{aligned}
    {\rm\mathbf{H}} = \frac{1}{\sqrt{\rho}}\left(\sqrt{\frac{\mathcal{K}}{\mathcal{K}+1}}\mathcal{L}_{LoS} + \sqrt{\frac{1}{\mathcal{K}+1}}\mathcal{L}_{NLoS}\right),
    \end{aligned}
\end{equation}
\endgroup
where $\mathcal{K}$ is the Rician factor, and $\mathcal{L}_{LoS}$ and $\mathcal{L}_{NLoS}$ are denoted as the LoS and NLoS components, respectively. The symbol $\rho$ refers to as the distance-based pathloss expressed as \cite{pathloss_reference} $PL\left(dB\right) = 10\log_{10}\rho = 32.4 + 20 \log_{10}\omega + 21 \log_{10}\varsigma$, where $\omega$ represents the central operating frequency and $\varsigma$ denotes the distance between the transmitter and receiver. Thereby, the effective STAR-RIS channel ${\boldsymbol\Omega_{i, k}}$ between the BS and $k$-th user in $i$-th region can be formulated as
\begingroup
\begin{subequations} \label{eqn: channel}
\begin{align}
    \label{channel}
    &{\boldsymbol\Omega_{i, k}} = {\rm\mathbf{d}}_{i, k}^{\mathcal{H}} \nonumber \\ 
        &+ \sum_{j \in \mathcal{V}}\sum_{j' \in \mathcal{C}'_{j}}\left(\mathbbm{1}_{\left\{C_{j,j',j} < i\right\}}\cdot{\rm\mathbf{g}}_{C_{j,j',j}, k}^{\mathcal{H}}{\boldsymbol{\Theta}}_{C_{j,j',j}}^{\mathcal{T}}f_{\prod}\left({\mathbf{C}}_{j,j'}\right) \right. \nonumber \\ 
        &\qquad\quad  \left. + \mathbbm{1}_{\left\{C_{j,j',j} \ge i\right\}}\cdot{\rm\mathbf{f}}_{ C_{j,j',j}, k}^{\mathcal{H}}
        {\boldsymbol{\Theta}}_{C_{j,j',j}}^{\mathcal{R}}f_{\prod}\left({\mathbf{C}}_{j,j'}\right)\right), \\
    \label{prod}
    &f_{\prod}\left({\mathbf{C}}_{j,j'}\right) = 
    \prod_{l=1}^{j-1}\left({\rm\mathbf{\Phi}}_{C_{j,j',l},C_{j,j',l+1}}
    {\boldsymbol{\Theta}}_{C_{j,j',l}}^{\mathcal{T}}\right){\rm\mathbf{\Phi}}_{C_{j,j',1}},
\end{align}
\end{subequations}
\endgroup
where $\mathbbm{1}$ denotes the indicator function representing the event occurrence and ${\mathcal{H}}$ represents the hermitian of a matrix. The function $\mathbbm{1}$ takes a value of $1$ when the event occurs and $0$ otherwise.
In \eqref{channel}, apart from ${\rm\mathbf{d}}_{i, k}^{\mathcal{H}}$ represented as the direct link, the other links are categorized into transmitted links of ${\rm\mathbf{g}}_{C_{j,j',j}, k}^{\mathcal{H}}{\boldsymbol{\Theta}}_{C_{j,j',j}}^{\mathcal{T}}f_{\prod}\left({\mathbf{C}}_{j,j'}\right)$ and reflected links of ${\rm\mathbf{f}}_{ C_{j,j',j}, k}^{\mathcal{H}}{\boldsymbol{\Theta}}_{C_{j,j',j}}^{\mathcal{R}}f_{\prod}\left({\mathbf{C}}_{j,j'}\right)$ based on a comparison of the indices of STAR-RISs $C_{j,j',j}$ and region $i$.
The product $f_{\prod}\left({\mathbf{C}}_{j,j'}\right)$ in \eqref{prod} can be used to calculate the cascaded STAR-RIS channel.

\subsection{Signal Model}
%\addtolength{\topmargin}{0.01in}
The transmitted signal and the active beamforming vector of the $k$-th user in the $i$-th region are defined as ${x}_{{i, k}} \in \mathbb{C}^{1 \times 1}$ and ${\rm\mathbf{w}}_{{i, k}} \in \mathbb{C}^{M \times 1}$, respectively.
In consequence, the received signal of the $k$-th user in the $i$-th region is given by
\begingroup
\allowdisplaybreaks
\begin{equation}\label{eqn: received signal}
\begin{aligned}
    y_{i, k} = {\boldsymbol\Omega_{i, k}} \sum_{i' \in \mathcal{I}}\sum_{k' \in \mathcal{K}_{i}}{\rm\mathbf{w}}_{i', k'}x_{i', k'} + n_{i, k}, 
\end{aligned}
\end{equation}
\endgroup
where $n_{i, k}$ is Gaussian noise. 
As known in \eqref{eqn: channel} and \eqref{eqn: received signal}, the corresponding SINR is given by
\begingroup
\allowdisplaybreaks
\begin{align} \label{eqn: SINR}
    &\Gamma_{i, k} = \nonumber \\ 
    &\frac{\left|{\boldsymbol{\Omega}}_{i, k}{\rm\mathbf{w}}_{i, k}\right|^{2}}{\sum_{k'\in \tilde{{\mathcal{K}}_i}}\left|{\boldsymbol{\Omega}}_{i, k}{\rm\mathbf{w}}_{i, k'}\right|^{2}+\sum_{i' \in \tilde{\mathcal{I}}}\sum_{k' \in {\mathcal{K}}_{i'}}\left|{\boldsymbol{\Omega}}_{i, k}{\rm\mathbf{w}}_{i', k'}\right|^{2} + \sigma^{2}},
\end{align}
\endgroup
where $\tilde{{\mathcal{K}}_i} = \left\{ \mathcal{K}_i \setminus k \right\}$ and $\tilde{\mathcal{I}} = \left\{ \mathcal{I} \setminus i \right\}$.
The first term in the denominator is the interference from users within the same region $i$, the second term is from users in the other regions, and $\sigma^{2}$ represents the noise power. According to \eqref{eqn: SINR}, the data rate of the system can be calculated as
\begingroup
\allowdisplaybreaks
\begin{equation}\label{eqn:data_rate}
    \begin{aligned}
    R = \sum_{i \in \mathcal{I}}\sum_{k \in \mathcal{K}_{i}}B\log_{2}\left(1 + \Gamma_{i, k}\right),
    \end{aligned}
\end{equation}
\endgroup
where $B$ represents the bandwidth.

% Power
Moreover, we denote the set of active beamforming as ${\rm\mathbf{W}} = \left\{{\rm\mathbf{W}}_{i}, \forall i \in \mathcal{I}\right\}$ with the beamforming matrix of the $i$-region ${\rm\mathbf{W}}_{i} = \left[{\rm\mathbf{w}}_{i, 1},...,{\rm\mathbf{w}}_{i, K_{i}}\right]$. Consequently, the power constraint for the BS is expressed as
\begingroup
\allowdisplaybreaks
\begin{equation} \label{txpower_constraint}
    \begin{aligned}
    &\sum_{i \in \mathcal{I}} \text{tr}\left\{{\rm\mathbf{W}}_{i}{\rm\mathbf{W}}_{i}^{\mathcal{H}} \right\} \leq P_{\text{max}}, 
    \end{aligned}
\end{equation}
\endgroup
where $\text{tr}\left\{\cdot\right\}$ denotes the trace operation of the matrix and $P_{\text{max}}$ is the transmit power budget for the BS. Moreover, the power of the $n$-th element of the $v$-th STAR-RIS is given by
\begingroup
\allowdisplaybreaks
\begin{equation} \label{constraint of alpha}
    \begin{aligned}
    &\gamma_{v, n} = p \cdot \alpha_{v, n}, \ \forall v \in \mathcal{V}, n \in \mathcal{N},
    \end{aligned}    
\end{equation}
\endgroup
where the constant $p$ represents the power consumed by each element of STAR-RISs. 
The total power consumption of the entire system is calculated as
\begingroup
\allowdisplaybreaks
\begin{equation} \label{power consumption}
    \begin{aligned}
    &{P}_{total} = \sum_{v \in \mathcal{V}}\sum_{n \in \mathcal{N}}\gamma_{v, n} + 
    \sum_{i \in \mathcal{I}} \text{tr}\left\{{\rm\mathbf{W}}_{i}{\rm\mathbf{W}}_{i}^{\mathcal{H}} \right\}. \end{aligned}   
\end{equation}
\endgroup
% Energy efficiency
Therefore, the total energy efficiency is given by
\begingroup
\allowdisplaybreaks
\begin{equation} \label{Energy Efficiency}
    \begin{aligned}
    &{E}\left({{\boldsymbol{\Lambda}}, {\boldsymbol{\Xi}}, {\rm\mathbf{W}}}\right) = \frac{R}{{P}_{total}},
    \end{aligned}    
\end{equation}
\endgroup
with the variables ${\boldsymbol{\Lambda}}, {\boldsymbol{\Xi}}$, and ${\rm\mathbf{W}}$.
\subsection{Problem Formulation}
The objective is to maximize the energy efficiency while guaranteeing the constraints of the STAR-RIS. The corresponding optimization problem is formulated as
\begingroup
\allowdisplaybreaks
\begin{subequations} \label{problem}
\begin{align}    
    &\mathop{\max}_{{\boldsymbol{\Lambda}}, {\boldsymbol{\Xi}}, {\rm\mathbf{W}}}\ \ {E}\left({{\boldsymbol{\Lambda}}, {\boldsymbol{\Xi}}, {\rm\mathbf{W}}}\right), \\
    \label{constrain beamforming}
    & \ \ \text{s.t.}\ \ \ \sum_{i \in \mathcal{I}} \text{tr}\left\{{\rm\mathbf{W}}_{i}{\rm\mathbf{W}}_{i}^{\mathcal{H}} \right\} \leq P_{\text{max}}, \\
    \label{couple beta}
    &\qquad \ \ \beta_{v, n}^{\mathcal{T}} = \sqrt{1-\left(\beta_{v, n}^{\mathcal{R}}\right)^2} , \qquad \forall v \in \mathcal{V}, n \in \mathcal{N}, \\
    \label{couple theta}
    &\qquad \ \ \theta_{v, n}^{\mathcal{T}} = 
    \theta_{v, n}^{\mathcal{R}} \pm \frac{\pi}{2}, \qquad \forall v \in \mathcal{V}, n \in \mathcal{N}, \\
    \label{constrain alpha}
    &\qquad \ \ \alpha_{v, n} \in \left\{0, 1\right\} , \qquad \forall v \in \mathcal{V}, n \in \mathcal{N}, \\
    \label{constrain beta}
    &\qquad \ \ 0 \leq {\beta_{v,  n}^{u}} \leq 1 , \qquad \forall v \in \mathcal{V}, n \in \mathcal{N}, {u} \in \mathcal{U}, \\
    \label{constrain theta}
    &\qquad \ \ 0 \leq {\theta_{v,  n}^{u}} \leq 2\pi , \qquad \forall v \in \mathcal{V}, n \in \mathcal{N}, {u} \in \mathcal{U}.
\end{align}
\end{subequations}
\endgroup
Constraint \eqref{constrain beamforming} is the maximum transmit power constraint of the BS. Constraints on amplitude in \eqref{couple beta} and phase shift in \eqref{couple theta} describe the coupling relationship between transmission and reflection capability of STAR-RISs. Constraints \eqref{constrain alpha}, \eqref{constrain beta} and \eqref{constrain theta} characterize the binary on-off state and limitations of the amplitude and phase shift of STAR-RISs. Problem \eqref{problem} presents an obstacle due to its non-convex and non-linear nature with fractional functions and logarithmic expressions. Therefore, we propose a DRL-based algorithm to deal with the above problems, which is elaborated in the following section.

\section{Proposed MAGAR Algorithm}
\begin{figure}
 \centering
 \includegraphics[width=3.2in]{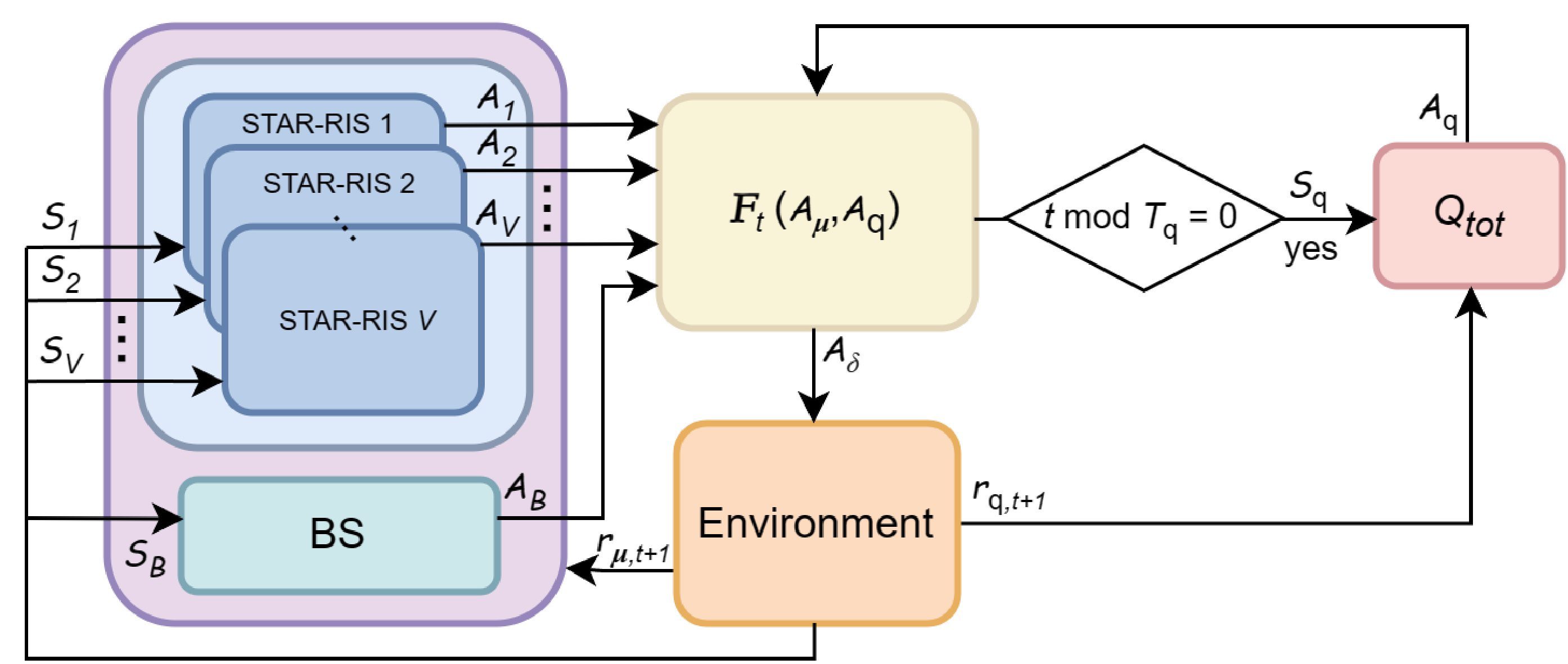} % change to eps file when this paper finish !
 \captionsetup{font=footnotesize}
 \caption{Proposed MAGAR algorithm.}
 \label{Fig:Algorithm_figure}
\end{figure}
However, a centralized solution employing a single-agent RL with large and complex observation spaces lead to insufficient memory storage and slow convergence. As a remedy, the multi-agent DRL (MA-DRL) algorithm enable each agent to operate independently, improving flexibility and efficiency in handling complex scenarios. However, training the independent agents for optimizing the team reward presents a challenge.
Motivated by these challenges, we propose MAGAR solves the complex problem \eqref{problem} by exchanging the execution of global and local agents. MAGAR not only effectively optimizes the global reward, but also enhances cooperation among local agents.
%DRL generally refers to the fundamental method of DQN, which leverages two neural networks, with one referred to as the \textit{current} network evaluating the action-value function, and the other as the \textit{target} network maintaining an identical structure to the current network to stabilize the learning process. In addition, DQN incorporates an experience replay mechanism, which stores the past experiences and then samples them uniformly during the training process. The proposed MAGAR solves the complex problem \eqref{problem} by exchanging the execution of global and local agents. MAGAR not only effectively optimizes the global reward, but also enhances cooperation among local agents.
\subsection{Markov Decision Process}
%A standard tuple representing a Markov decision process (MDP) is denoted as $(\mathcal{S}_{t}, \mathcal{A}_{t}, P_{t}, r_{t+1})$, where $\mathcal{S}_{t}$ is the state space and $\mathcal{A}_{t}$ is the action space at time slot $t$. The transition function $P_{t}$ indicates the probability distribution over states after performing an action, and $r_{t+1}$ at time slot $t$ denotes the reward function calculated when taking an action in a given state. 
By reformulating \eqref{problem} as a Markov decision process (MDP), our objective is to maximize the accumulated reward $\mathbb{E}[\sum_{t=1}^{T}\gamma^{t} r_{t+1}]$, where $\gamma \in [0, 1]$ denotes the discount factor, $r_{t+1}$ is the reward function and $T$ is the end time slot of an episode.
We define all STAR-RISs and BS as agents with their own MDP $(\mathcal{S}_{\mu,t}, \mathcal{A}_{\mu,t}, P_{\mu,t}, r_{\mu, t+1})$, $\forall \mu \in \left\{v,{\mathcal{B}}, \forall v \in \mathcal{V}\right\}$ and ${\mathcal{B}}$ represents the agent of the BS. Moreover, we define $(\mathcal{S}_{{\rm q},t}, \mathcal{A}_{{\rm q},t}, P_{{\rm q},t}, r_{{\rm q}, t+1})$ as the MDP of the global agent $Q_{tot}$, where the reward $r_{{\rm q},t}$ is shared among the whole system. In the following, we simplify the sets of states and actions $\mathcal{S}_{\mu,t}$, $\mathcal{S}_{{\rm q},t}$, $\mathcal{A}_{\mu,t}$, and $\mathcal{A}_{{\rm q},t}$ as $\mathcal{S}_{\mu}$, $\mathcal{S}_{\rm q}$, $\mathcal{A}_{\mu}$, and $\mathcal{A}_{\rm q}$, respectively.\\

\subsection{Proposed MAGAR Algorithm}
\label{section:MAGAR}
As depicted in Fig. \ref{Fig:Algorithm_figure}, our proposed MAGAR algorithm comprises two components. The first component involves independent training for each agent to interact with the environment and receives an agent-specific reward $r_{\mu,t}$. This promotes individual exploration where agents may converge to the suboptimal actions due to limited observation. 
The second component emphasizes global optimization of the entire system by a global agent denoted as $Q_{tot}$.
Two different actions $\mathcal{A}_{\mu}$ for each agent and $\mathcal{A}_{{\rm q}}$ for the global agent will be performed based on
\begingroup
\allowdisplaybreaks
\begin{align}
    &F_{t}\left(\mathcal{A}_{\mu}, \mathcal{A}_{{\rm q}}\right) = \left\{
    \begin{aligned}
      &\mathcal{A}_{{\rm q}},\ \text{if}\ t\ \text{mod}\ {T_{\rm q}} = 0, \\
      &\mathcal{A}_{\mu},\ \text{otherwise},
    \end{aligned}
    \right.
\end{align}
\endgroup
where $T_{{\rm q}} > 0$ represents the execution period by the global agent. 
With comprehensive information gathered from all agents within the global state $\mathcal{S}_{{\rm q}}$, the global agent is empowered to make better decisions $\mathcal{A}_{{\rm q}}$. Although $Q_{tot}$ can acquire complete state information from all agents, frequent execution of the global agent leads to computational overhead.
Therefore, the periodic nature of the replacement by $Q_{tot}$ strikes a balance between achieving global optimization and avoiding burdening the system.

\begin{itemize}
    \item \textbf{\textit{State}}: For each STAR-RIS agent, we denote $\mathcal{S}_{v} = \left\{{\beta_{v, n}^{\mathcal{R}}}, {\theta_{v, n}^{\mathcal{R}}}, {\theta_{v, n}^{\mathcal{T}}}, {\alpha_{v, n}}, \forall n \in \mathcal{N}\right\}$ as the state of the $v$-th STAR-RIS, comprising the amplitude of part ${\mathcal{R}}$, phase shifts of both parts ${\mathcal{R}}$ and ${\mathcal{T}}$, as well as on-off state. 
    For the BS agent, we decompose the beamforming matrix into real and imaginary parts, as the state of $\mathcal{S}_{\mathcal{B}} = \left\{{{\rm\mathbf{W}}^{\zeta}}, {{\rm\mathbf{W}}^{\eta}}\right\}$, where ${\zeta}$ and ${\eta}$ are the real and imaginary operation.
    For the global agent, we denote $\mathcal{S}_{\rm q} = \left\{\mathcal{S}_{v}, \mathcal{S}_{\mathcal{B}}, \forall v \in \mathcal{V} \right\}$.
    %In addition, we define a general state $\mathcal{S}_{\delta}, \forall \delta \in \left\{\mu, {\rm q}\right\}$ as the input of the environment.
    \item \textbf{\textit{Action}}: For each STAR-RIS agent, we define the action set which includes increment/decrement of amplitude and phase shifts, as well as on-off policy as $\mathcal{A}_{v} = \left\{\pm{\Delta_{\beta^{\mathcal{R}}_{v,n}}}, \pm{\Delta_{\theta^{\mathcal{R}}_{v,n}}}, {\vartheta_{v,n}}, {\alpha}_{v,n}, \forall v \in \mathcal{V}, n \in \mathcal{N}\right\}$, where ${\vartheta_n}$ determines the phase difference between $+\frac{\pi}{2}$ and $-\frac{\pi}{2}$ as well as ${\alpha}_{n} \in \left\{0, 1\right\}$. Here, ${\Delta_{\beta^{\mathcal{R}}_{v,n}}} > 0$ and ${\Delta_{\theta^{\mathcal{R}}_{v,n}}} > 0$ represent redefined step sizes for ${\beta_{v, n}^{\mathcal{R}}}$ and ${\theta_{v, n}^{\mathcal{R}}}$ within the intervals $[0, 1]$ and $[0, 2\pi)$, respectively. 
    For the BS agent, we define the set of actions for increment/decrement of the real and imaginary parts of the beamforming matrix as $\mathcal{A}_{\mathcal{B}} = \left\{\pm{\Delta_{\zeta}}, \pm{\Delta_{\eta}}\right\}$, where $\Delta_{\zeta} > 0$ and $\Delta_{\eta} > 0$ represent the step size of each BS's antenna. 
    For the global agent, we denote $\mathcal{A}_{\rm q} = \left\{\mathcal{A}_{v}, \mathcal{A}_{\mathcal{B}}, \forall v \in \mathcal{V} \right\}$.
    %Additionally, we define a general action $\mathcal{A}_{\delta}, \forall \delta \in \left\{\mu, {\rm q}\right\}$ as the output of the environment.
\end{itemize}

We denote $s_{\delta,t} \in \mathcal{S}_{\delta}$
and $a_{\delta,t} \in \mathcal{A}_{\delta}$ as the input and output of the environment.
At the $t$-th time slot, an agent select an action $a_{\delta,t} \in \mathcal{A}_{\delta}$ based on the $\epsilon$-greedy policy as
\begingroup
%\allowdisplaybreaks
\begin{align}
    a_{\delta,t} = \left\{
    \begin{aligned}
      & \text{Randomly select from}\ \mathcal{A}_{\delta},\quad \ \text{if}\ rand() \leq \epsilon, \\
      & \text{argmax}_{a_{\delta,t}}Q\left(s_{\delta,t}, a_{\delta,t}\right),\quad\quad\ \text{otherwise},
    \end{aligned}
    \right.
\end{align}
\endgroup
with the exploration probability defined as $\epsilon$, the state as $s_{\delta,t} \in \mathcal{S}_{\delta}$, and the $Q\left(s_{\delta,t}, a_{\delta,t}\right)$ as action-value function.
The next state will be updated as $s_{\delta,t+1} \leftarrow s_{\delta,t} + a_{\delta,t}$ if the value is within the legitimate state space. Otherwise, the next state will remain unchanged as $s_{\delta,t+1} \leftarrow s_{\delta,t}$. Furthermore, to satisfy the constraint \eqref{constrain beamforming}, we normalize the beamforming matrix by $\sum_{i \in \mathcal{I}} \text{tr}\left\{{\rm\mathbf{W}}_{i}{\rm\mathbf{W}}_{i}^{\mathcal{H}} \right\} = P_{\text{max}}$. 
With the constraint of STAR-RIS \eqref{constrain beta}, $\beta_{v, n}^{\mathcal{T}}$ and ${\boldsymbol{\Theta}_{v}^{u}}$ could be obtained.
Therefore, at $t$-th time slot, the updated state of each agent is employed to compute the individual reward $R_{\mu,t}$ and power consumption ${P}_{\mu, t}$. In particular, except the states of agent $i$ all agents remain unaltered during the calculation of $R_{\mu,t}$.

%, where $\forall \delta \in \left\{\mu, {\rm q}\right\}$.

Moreover, we define the global reward function as 
\begingroup
\allowdisplaybreaks
\begin{align}
    \label{reward: all}
    r_{{\rm q}, t} = \frac{R_{t}}{{P}_{total, t}},
\end{align}
\endgroup
where $R_{t}$ and ${P}_{total, t}$ represent the achievable rate and the total power consumption obtained at $t$-th time slot.
However, as observed in \cite{lazy_agent_reference}, MA-RL may have the issue of partial local observability. A few agents learn a beneficial policy, while the other agents intend to act lazily during learning owing to the shared global reward. Accordingly, we introduce an additional reward function for each agent as
\begingroup
\allowdisplaybreaks
\begin{align}
    \label{reward: individual}
    r_{\mu, t} = \frac{R_{\mu,t}}{{P}_{\mu, t}}. 
\end{align}
\endgroup

The optimal target value for each agent is indicated as $y_{\mu} = r_{\mu, t+1} + \gamma{ \max_{a_{\mu}}Q_{\mu, t} \left(s_{\mu, t+1}, a_{\mu}\right)}$. Therefore, agents learn their optimal policies through refining the individual Q-value function, which can be represented by 
\begingroup
%\allowdisplaybreaks
\begin{align}
    \label{Q update: individual}
    &Q_{\mu, t+1}\left(s_{\mu, t}, a_{\mu, t}\right) \gets \left(1 - \lambda\right) Q_{\mu, t}\left(s_{\mu, t}, a_{\mu, t}\right) + \lambda y_{\mu} .
\end{align}
\endgroup
Here, $\lambda$ denotes the learning rate. The notation $r_{\mu, t+1}$ is the immediate reward obtained by executing the action $a_{\mu, t} \in \left\{\mathcal{A}_{v}, \mathcal{A}_{\mathcal{B}}\right\}, \forall {v} \in \mathcal{V}$ in the state $s_{\mu, t} \in \left\{\mathcal{S}_{v}, \mathcal{S}_{\mathcal{B}}\right\}, \forall {v} \in \mathcal{V}$ at the $t$-th time slot. We employ a neural network function approximator with weights $\xi$ as a Q-network. Otherwise, for each agent, we define $\psi_{\mu} = r_{\mu, t+1} + \gamma{ \max_{a_{\mu}}Q_{\mu, t} \left(s_{\mu, t+1}, a_{\mu};\xi^{-}_{\mu,\tau}\right)}$, where $\xi^{-}_{\mu,\tau}$ denotes the Q-network parameters from some previous time slots. Consequently, the loss function can be given by
\begingroup
\begin{align} 
\label{loss: individual}
    L_{\mu} \left(\xi_{\mu,\tau}\right) = 
    \mathbb{E}\left[ \left(\psi_{\mu} - Q_{\mu, t}\left( s_{\mu, t}, a_{\mu, t};\xi_{\mu,\tau} \right) \right)^2\right], 
\end{align}
\endgroup
where $\xi_{\mu,\tau}$ represent the parameter at $\tau$ iteration of the current network for each agent.

However, limited observation potentially leading to the local optimum. To address this issue, we propose a cooperative multi-agent architecture to facilitate information exchange and coordination among agents. The complete information $s_{{\rm q}, t} \in \mathcal{S}_{{\rm q}} = \mathcal{S}_{v} \cup \mathcal{S}_{\mathcal{B}}$ is aggregated from each individual agent. Then, the global agent determines a joint action $a_{{\rm q}, t} \in \mathcal{A}_{{\rm q}} = \mathcal{A}_{v} \cup \mathcal{A}_{\mathcal{B}}$ from each agent. Moreover, we determine $\max_{a_{\mu}}Q_{\mu, t} \left(s_{\mu, t}, a_{\mu}\right)$ as the maximum Q value obtained from each agent in the previous time slot. 
For the global agent, we define the optimal target value as $y_{{\rm q}} = r_{{\rm q}, t+1} + \gamma\sum_{\mu}{\max_{a_{\mu}}Q_{\mu, t} \left(s_{\mu, t}, a_{\mu}\right)}$, which incorporates the immediate reward $r_{{\rm q}, t+1}$ predefined in \eqref{reward: all}.
Specifically, summarizing the maximum Q values from local agents elevates the cooperation of the entire system.
Therefore, we approximate the optimal action-value function of global agent by $Q_{q, t+1}\left(s_{{\rm q}, t}, a_{{\rm q}, t}\right)$, which is analogous to \eqref{Q update: individual} but with subscript $\mu$ replaced by ${\rm q}$. 
To train the global Q-network, we employ the parameters $\xi^{-}_{{\rm q},\tau}$ from the previous time slots, where the optimal target values $y_{{\rm q}}$ are substituted with approximate target values $\psi_{{\rm q}} = r_{{\rm q}, t+1} + \gamma\sum_{\mu}{\max_{a_{\mu}}Q_{\mu, t} \left(s_{\mu, t}, a_{\mu};\xi^{-}_{{\rm q},\tau}\right)}$.
This leads to a sequence of the loss function of $Q_{tot}$, $L_{\rm q}\left(\xi_{\rm q,\tau}\right)$, which is achieved by substituting the subscript symbol $\mu$ in \eqref{loss: individual} with ${\rm q}$.

\section{Performance Evaluation}
In this section, the results of the simulations are presented to evaluate the performance of the proposed MAGAR algorithm in multi-hop STAR-RIS system, serving total users $K=\sum_{i \in \mathcal{I}}K_i$. The transmit power of the BS is set to $33$ dBm, whilst the power of each STAR-RIS element is set to $p=17$ dBm. The noise power is $\sigma^{2} = -174 + 10\log_{10}B$ dBm, where $B=100$ MHz is the operating bandwidth. The carrier frequency is set to $\omega=28$ GHz.
In our MAGAR algorithm, the simulation parameters are as $\left\{T_{\rm{q}}, \epsilon, \lambda, \gamma \right\} = \left\{20, 0.3, 0.1, 0.9\right\}$.

\begin{figure}[!t]
 \centering
 \includegraphics[width=3in]{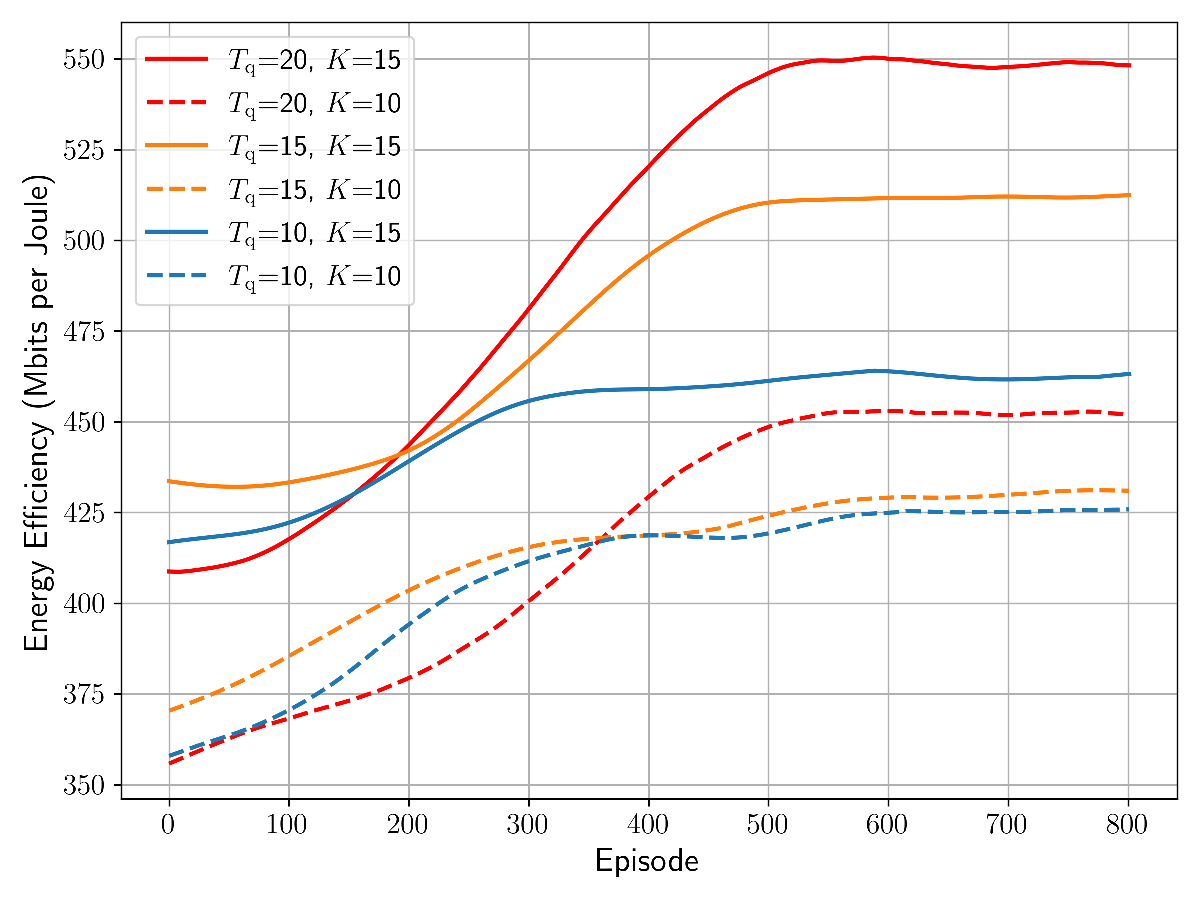} % change to eps file when this paper finish !
 \captionsetup{font=footnotesize}
 \caption{Convergence behavior of the proposed MAGAR algorithm.}
 \label{fig:converge}
\end{figure}

%\begin{figure*}
%	\centering
%\begin{minipage}[t]{0.24\textwidth}
%	\centering
% \includegraphics[width=1.6in]{fig/NVee_on_7.eps} % change to eps file when this paper finish !
% \captionsetup{font=footnotesize}
% \caption{The energy efficiency performance versus different numbers of $N$ and $V$ under different operating mechanism of all-on, half-on, and optimized.}
% \label{fig:NV_on}	
%\end{minipage}
%\
%\begin{minipage}[t]{0.24\textwidth}
%	\centering
%	\includegraphics[width=1.6in]{fig/repltK2_ee2.eps}
% \captionsetup{font=footnotesize}
% \caption{The energy efficiency of the system assisted by multi-hop STAR-RIS, STAR-RIS MS, RIS, and deployment without RIS.}
% \label{fig:4}
%\end{minipage}	
%\	
%\begin{minipage}[t]{0.24\textwidth}
%	\centering
% \includegraphics[width=1.6in]{fig/deployEE.eps}
% \captionsetup{font=footnotesize}
% \caption{Comparison of proposed MAGAR algorithm and benchmarks versus the various distance between STAR-RISs $\varsigma$.}
% \label{fig:5}
%\end{minipage}	
%\	
%\begin{minipage}[t]{0.24\textwidth}
%	\centering
% \includegraphics[width=1.6in]{fig/M_alg_ee.eps}
% \captionsetup{font=footnotesize}
% \caption{The energy efficiency of MAGAR and benchmarks versus different numbers of BS transmit antennas $M$.}
% \label{fig:6}
%\end{minipage}		
%\end{figure*}

In Fig. \ref{fig:converge}, we investigate the convergence of MAGAR across varying numbers of users, considering $M=5$, $V=3$, $N=16$, $\varsigma=10$. We observe that the frequency of the global optimization agent $Q_{tot}$ plays a crucial role in influencing reward performance. Infrequent execution fosters autonomous exploration by agents, potentially leading to effective reward optimization, while excessive execution of $Q_{tot}$ can introduce overhead and hinder reward acquisition. Moreover, the simulation results demonstrate a positive correlation between the number of users served by the system and the reward.

\begin{figure}[!t]
 \centering
 \includegraphics[width=3in]{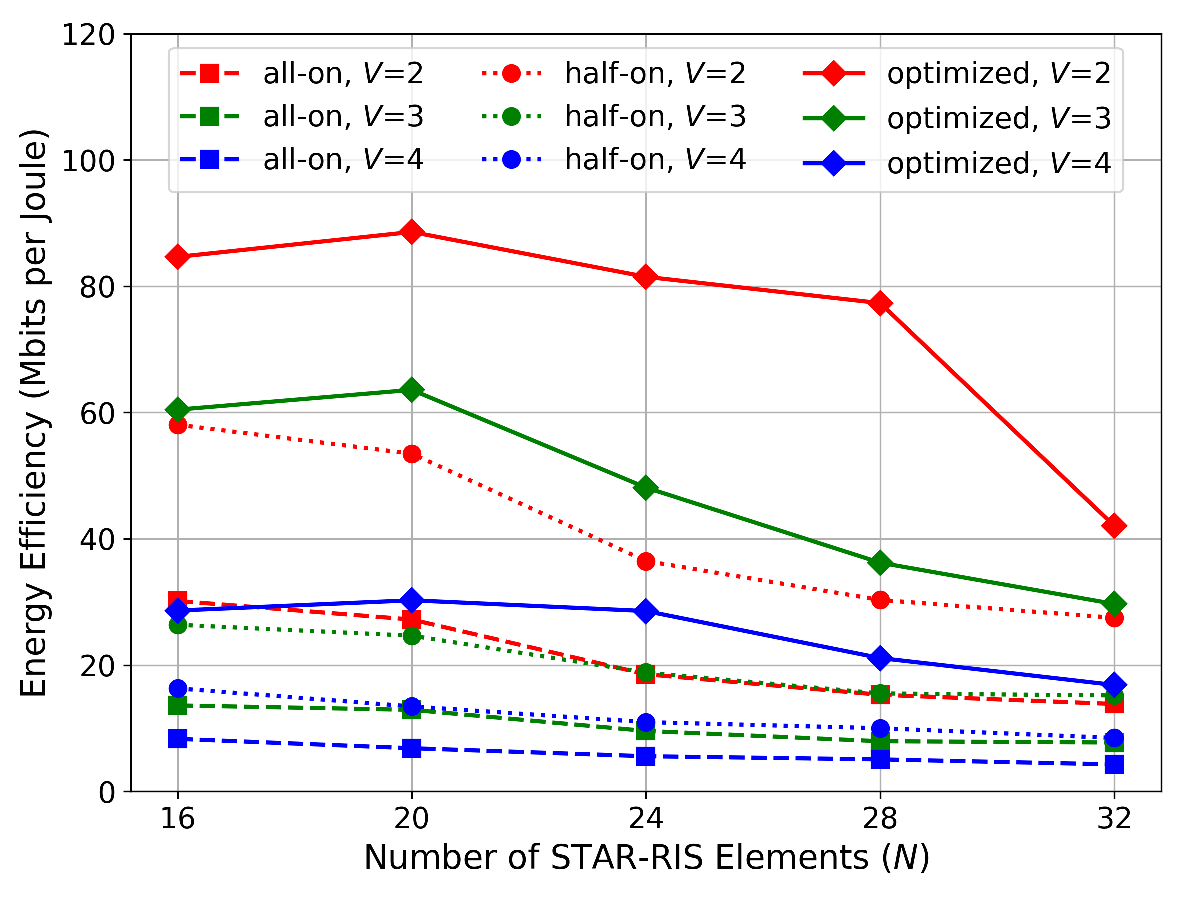} % change to eps file when this paper finish !
 \captionsetup{font=footnotesize}
 \caption{The energy efficiency performance versus different numbers of $N$ and $V$ under different operating mechanism of all-on, half-on, and optimized.}
 \label{fig:NV_on}
\end{figure}

In Fig. \ref{fig:NV_on}, we evaluate the performance of MAGAR under different numbers of STAR-RIS elements $N$, and number of STAR-RISs $V$. We consider $T_{\rm q}=20$, $M=5$, $K=10$, and $\varsigma=10$. Also, we investigate the performance of three on-off policies, \textbf{all-on}: all elements are switched on, \textbf{half-on}: randomly switch on half elements of the STAR-RISs, \textbf{optimized}: dynamically and optimally selected by MAGAR.
We observe that the trend in energy efficiency exhibits an initial ascent followed by a subsequent decline. As the number of elements continues to grow, the dominance of power consumption outweighs the gains in the sum rate, ultimately leading to a deterioration of energy efficiency. The results demonstrate that the optimized on-off mechanism outperforms the other two cases. Specifically, the configurations in which all elements are operational exhibit the worst performance, owing to the associated high power cost.

\begin{figure}[!t]
\centering
 \includegraphics[width=3in]{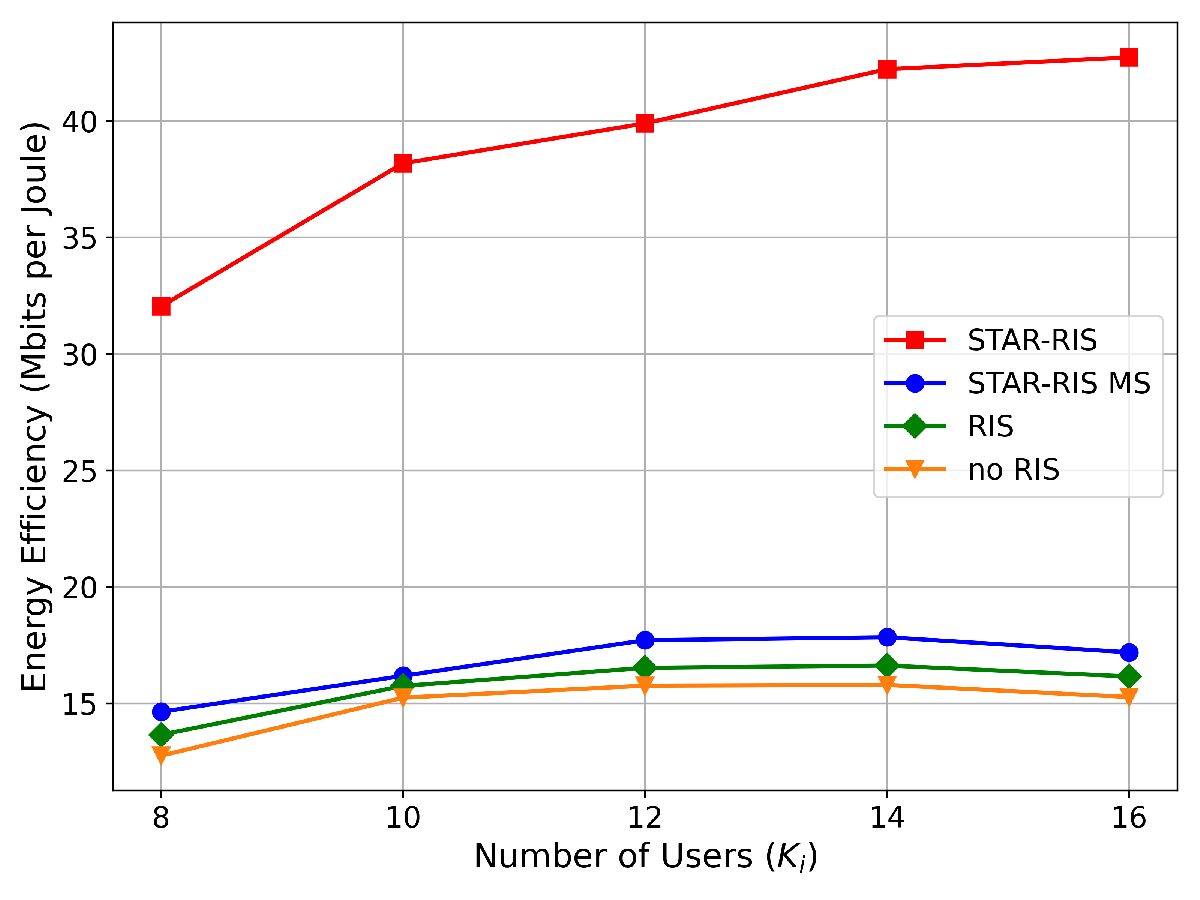}
 \captionsetup{font=footnotesize}
 \caption{The energy efficiency of the system assisted by multi-hop STAR-RIS, STAR-RIS MS, RIS, and deployment without RIS.}
 \label{fig:4}
\end{figure}

In Fig. \ref{fig:4}, we study the impact of users within the multi-hop STAR-RIS-aided system across various operational modes, including the ES mode, mode switching (MS) based STAR-RISs \cite{STAR-RIS_mode}, conventional RISs, and the scenario without RISs. 
We consider $T_{\rm q}=20$, $M=5$, $V=2$, $N=16$, and $\varsigma=10$.
We can observe that the augmenting of serving users correlates with increasings energy efficiency. In contrast to conventional RISs, STAR-RIS offers superior performance. Furthermore, the proposed ES-based multi-hop STAR-RISs demonstrate the highest EE performance. This is attributed to the amplitude control capabilities of all the STAR-RIS elements, which facilitate superior interference management compared to the fixed amplitude in MS mode. In contrast, transmission without RIS only provides direct link channels, resulting in inferior performance.

\begin{figure}[!t]
\centering
 \includegraphics[width=3in]{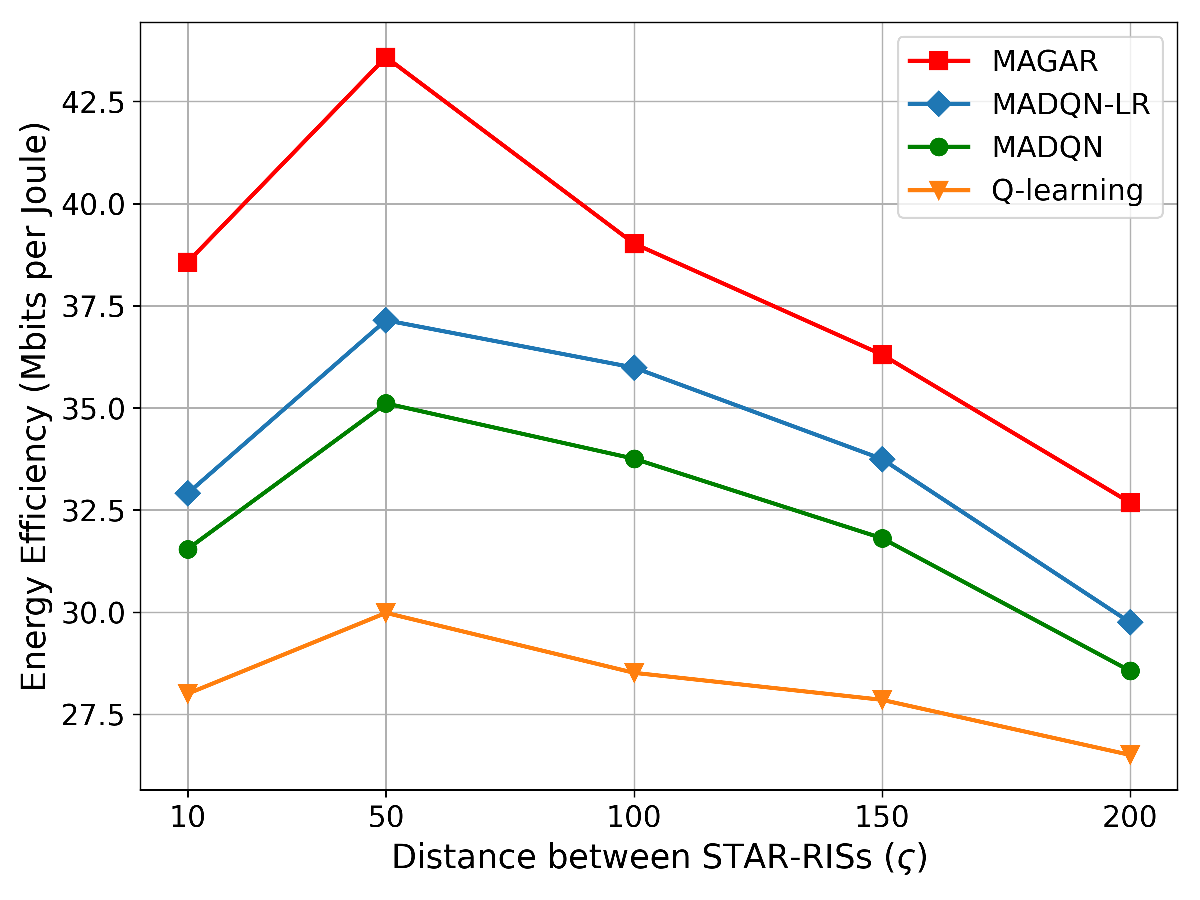}
 \captionsetup{font=footnotesize}
 \caption{Comparison of proposed MAGAR algorithm and benchmarks versus the various distance between STAR-RISs $\varsigma$.}
 \label{fig:5}
\end{figure}
The impact of STAR-RIS deployment is analyzed in Fig. \ref{fig:5}, with $T_{\rm q}=20$, $M=5$, $V=2$, $N=16$, and $K=10$. We consider the deployment of the STAR-RIS is along the x-axis with an inter STAR-RIS distance of $\varsigma$, where the first element is placed at $(0, \varsigma)$. Users are randomly distributed in the x-coordinate range of $0$ to $5\varsigma$ and the y-coordinate range of 0 to $\varsigma$. In \textbf{Q-learning}, \textbf{MADQN-LR}, and \textbf{MADQN}, we employ Q-learning \cite{Q_learning_reference}, Multi-agent DQN \cite{MADQN_reference} with global reward, and Multi-agent DQN utilizing local rewards for each agent, respectively. Within $\left[10,50\right]$ m, we observe an increased performance owing to the reduced channel correlation of the STAR-RISs and the shorter distance between STAR-RISs and users. However, performance declines when the distance $\varsigma$ extends from $50$ to $200$ m due to dominant high pathloss.

\begin{figure}[!t]
\centering
 \includegraphics[width=3in]{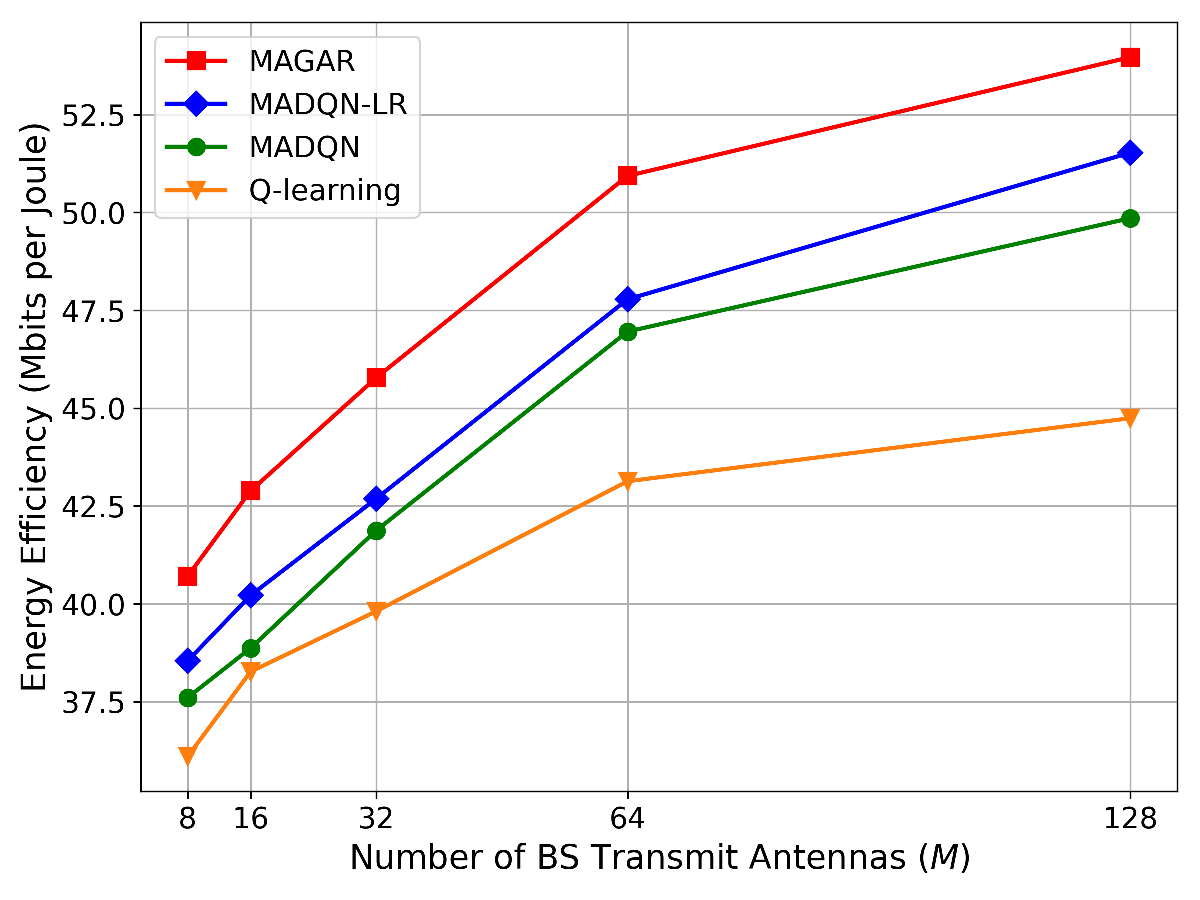} % change to eps file when this paper finish !
 \captionsetup{font=footnotesize}
 \caption{The energy efficiency of MAGAR and benchmarks versus different numbers of BS transmit antennas $M$.}
 \label{fig:6}
\end{figure}

In Fig. \ref{fig:6}, we compare the proposed MAGAR algorithm with several benchmarks with different numbers of BS transmit antennas $M$, cosidering $T_{\rm q}=20$, $V=2$, $N=16$, $K=10$, and $\varsigma=10$.
We observe that MADQN-LR with local rewards outperforms MADQN using global rewards with around $3\%$. The proposed MAGAR algorithm, incorporating the global agent, exhibits a significant enhancement in energy efficiency with $7\%$. As for Q-learning, higher number of BS transmit antennas leads to little improvement of energy efficiency diminishes, as it lacks the neural networks to handle a comparatively large action space under complex environments.

\section{Conclusion}
In this paper, we have conceived an multi-hop STAR-RIS-assisted transmissions, aiming for maximizing the energy efficiency. We develop a MAGAR algorithm to jointly optimize the active BS beamforming and configurations of STAR-RISs considering the on-off state of STAR-RIS elements. Simulations have demonstrated the effectiveness of the proposed MAGAR scheme in terms of different numbers of STAR-RISs, STAR-RIS elements, transmit antennas, and users. Furthermore, benefited from global agent enhancing collaboration of local agents, our proposed MAGAR algorithm achieves the highest energy efficiency compared to other existing benchmarks.

%\linespread{0.86}
\bibliographystyle{IEEEtran}

\bibliography{reference}

\end{document}